\documentclass[sigconf,authoryear]{acmart}

\usepackage{graphicx}
\usepackage{booktabs}
\copyrightyear{2026}
\acmYear{2026}
\setcopyright{cc}
\setcctype{by}
\acmConference[SA Technical Communications '26]{SIGGRAPH Asia 2026 Technical Communications}{December 01--04, 2026}{Kuala Lumpur, Malaysia}
\acmBooktitle{SIGGRAPH Asia 2026 Technical Communications (SA Technical Communications '26), December 01--04, 2026, Kuala Lumpur, Malaysia}
\acmDOI{10.1145/3829339.3847852}
\acmISBN{979-8-4007-2841-9/2026/12}

\begin{document}

\title{The Weight Is Over --- Interactive Diffusion on Consumer GPUs}

\author{Frieder Ganz}
\orcid{0000-0002-6140-4805}
\affiliation{%
  \institution{Adobe}
  \city{Hamburg}
  \country{Germany}}
\email{ganz@adobe.com}

\author{Maximilian Müller}
\orcid{0000-0001-7424-9712}
\affiliation{%
  \institution{NVIDIA}
  \city{W\"urselen}
  \country{Germany}}
\email{maximilianm@nvidia.com}

\renewcommand{\shortauthors}{Ganz and Müller}

\begin{abstract}
On-device inference is booming, but the momentum is almost all in language models. Deploying image generators on consumer hardware in production remains hard: diffusion pipelines are memory-hungry, latency-sensitive, and require orchestrating an embedder, a transformer, a decoder, and often further postprocessing that is not as standardized as LLM inference loops are. We navigate the trade-off between performance, quality, and model footprint to reach as many client devices in the wild as possible. Central is low-bit quantization (FP8/NVFP4), which yields two separable effects: weight quantization gives a significant reduction in memory footprint, while activation quantization adds a speedup on compute-bound stages. For multi-model pipelines whose weights exceed VRAM, a carefully designed GPU-memory offloading scheme is required across text encoders, VAE, and auxiliary conditioning models. The result is an interactive on-device diffusion image generation editor with a time-to-first-image-iteration (TTFI) under one second on recent GPUs, and maintained quality of service across a much wider set of hardware.

We make three contributions: (1) an embedding translator that maps a small text encoder into a large encoder's space to cut weight and latency; (2) a reproducible sweep recipe for navigating the speed/quality/memory triangle in diffusion pipelines; and (3) an interactive on-device image generation editor achieving sub-second TTFI on recent GPUs.
\end{abstract}



\begin{teaserfigure}
  \centering
  \includegraphics[width=0.76\textwidth]{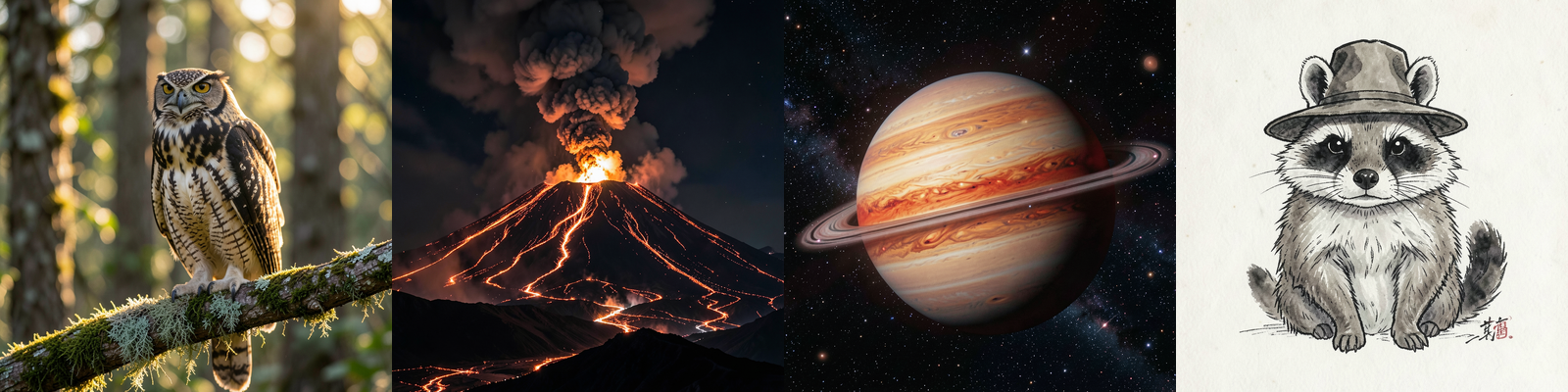}
  \caption{FLUX.2-klein with its 4B text encoder replaced by Qwen3-0.6B plus a 187M-parameter \emph{translator} (trained post-hoc; transformer and VAE frozen). Prompts: a horned owl at golden hour; an erupting volcano at night; Jupiter; a sumi-e ink-wash raccoon.}
  \Description{Four generated images: an owl, an erupting volcano, the planet Jupiter, and an ink-wash anthropomorphic raccoon.}
  \label{fig:teaser}
\end{teaserfigure}

\maketitle

\section{Introduction}
Diffusion image generation on consumer hardware requires re-imple\-mentation of the research code in a hardware-agnostic manner for large-scale deployment on client devices. Debugging precision and pipeline flow across text encoder, denoising transformer, decoder, and auxiliary conditioning has many difficulties. All models must fit in limited video memory and run within an interactive latency budget. This ideally works on a substantial fraction of client devices, and not just on the largest GPU. We treat this as a speed/quality/memory triangle and attack it on three fronts.
We focus on the first contribution, the embedding translator (Section~\ref{sec:translator}); the quantization and offloading recipe (Section~\ref{sec:system}) and the interactive image generation editor (Section~\ref{sec:editor}) are summarized.

\section{State of the Art}
Text-to-image diffusion has moved from U-Net latent models~\cite{rombach2022ldm} to rectified-flow transformers~\cite{esser2024sd3}, while sampling has been compressed to a few steps and deployment pushed onto mobile and consumer GPUs~\cite{li2023snapfusion,zhao2024mobilediffusion}. These systems still condition on large pretrained text encoders; we instead ask how small that encoder can be, aligning a compact encoder to a large one's features in the spirit of knowledge distillation~\cite{hinton2015distilling}; concurrent work likewise scales down diffusion text encoders~\cite{wang2025scalingdown}. Closest to us are adapter bridges that connect a frozen text encoder to a frozen diffusion model: ELLA attaches a large LLM through a timestep-aware connector~\cite{hu2024ella}, and LaVi-Bridge couples arbitrary language and vision backbones via LoRA plus an adapter~\cite{zhao2024lavibridge}. Both add capacity to improve prompt following; our translator instead runs this mapping in reverse, aligning a much \emph{smaller} encoder to the large one's exact conditioning space while leaving the diffusion model untouched, trading a little quality for large memory and latency savings on-device.

\section{Embedding Translator}
\label{sec:translator}
Diffusion models are usually trained with large text embedders taken from language models such as Qwen. We argue that for simple prompts, much smaller embeddings are enough. We propose a \emph{translator} network that sits between a small text embedder and the diffusion model and maps the small embedder's features into the conditioning the original large embedder produced. It is trained after the main pipeline is done---the diffusion model and VAE are left frozen---so it is a small, post-hoc add-on rather than a retraining step. Beyond just swapping in a smaller embedder, this lets us reuse an embedder that is \emph{already on the device}: for instance, the text features of an OS-shipped foundation model such as Apple's on-device LLM, so the image model ships no text encoder of its own. Translator networks can be of varying size (see our ablation, Table~\ref{tab:ablation}), but are in general far smaller than the original embedder. Each translator is trained post-hoc to regress the 0.6B encoder's per-token features onto the frozen 4B conditioning under a feature-regression loss (the diffusion transformer and VAE stay frozen); variants are named by hidden width and depth---e.g., d3072 is a single 3072-wide MLP and d2048$\times$8 eight 2048-wide blocks, while \emph{attn} rows use attention blocks.

We study this on FLUX.2-klein~\cite{flux2024}, replacing its 4B Qwen3~\cite{qwen3} encoder with Qwen3-0.6B plus a translator. Because all Qwen3 models share a tokenizer, the tokens line up and the translator can act token by token. We swept translator sizes from 3.5M to 187M parameters and measured image quality end to end on PartiPrompts~\cite{yu2022parti} (LPIPS~\cite{zhang2018lpips} and CLIPScore~\cite{hessel2021clipscore} against 4B images at the same seed), all at $1024\times1024$ (1K) resolution. Since translator outputs are not spatially aligned to the 4B reference, we treat LPIPS as a secondary deviation measure and CLIPScore as the primary semantic signal. How closely the translator matches the 4B features turned out not to predict image quality---that similarity stays flat at 0.73--0.77 for every size, while LPIPS ranges from 0.514 to 0.717. What matters is capacity: bigger translators steadily give better images. The largest (187M) reaches an LPIPS of 0.514---below the 0.553 we get just from re-running the 4B with a different seed---and a CLIPScore of 0.306, close to the 4B's 0.328; the smallest (3.5M) is clearly worse. The cost stays low and roughly the same at any translator size: the 0.6B pass takes 88\,ms and the translator adds at most 19\,ms, so encoding runs in about 90--107\,ms versus 435\,ms for the 4B (PyTorch/MPS on a Mac M3~Max, 48\,GB; in the deployed TensorRT-RTX pipeline the encoder runs in 4--12\,ms and is off the critical path (Table~\ref{tab:blackwell}), so on-GPU the translator's gain is chiefly memory). The encoder-path weight memory drops from 8\,GB to 1.4--1.8\,GB (Table~\ref{tab:ablation}) and is dominated by the shared 0.6B base, not the translator, so translator size trades quality for negligible memory. Attention translators did not beat a same-size MLP. Two caveats, both consistent with the ``simple prompts'' claim: the best translator still follows prompts slightly worse than the 4B (0.306 vs 0.328), and no configuration---the 4B included---renders readable text, which is the 4-step sampler's doing, not the encoder's.

\begin{figure}
  \includegraphics[width=0.85\columnwidth]{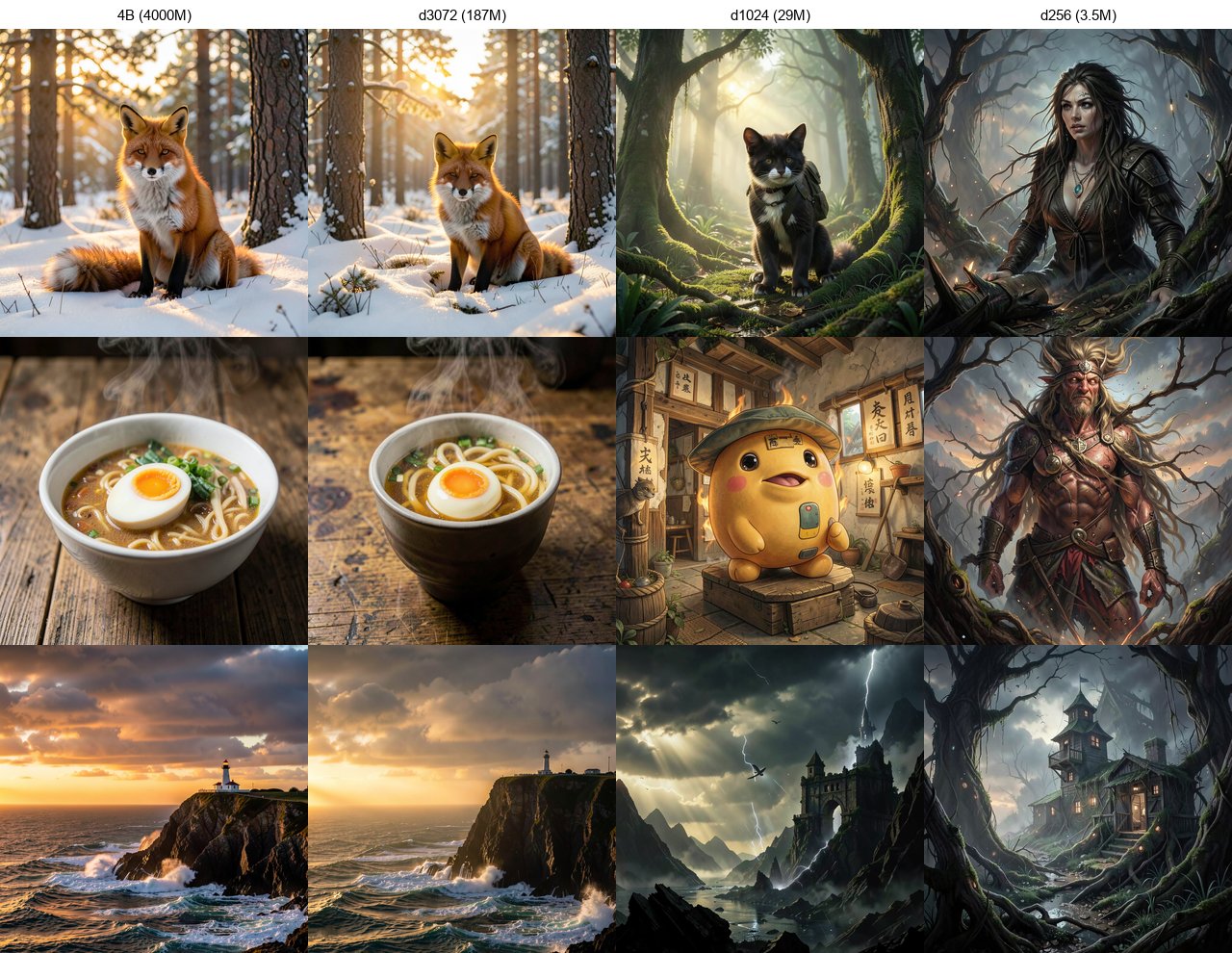}
  \caption{Same prompt and seed across encoders (columns): the 4B reference, and Qwen3-0.6B with a 187M, 29M, and 3.5M translator. The 187M translator tracks the 4B; the 29M drifts to plausible but wrong content; and the 3.5M collapses toward a single dominant style regardless of the prompt. Rows: a red fox in a snowy forest; a bowl of ramen; a lighthouse on a cliff at sunset. All images are $1024\times1024$ (1K).}
  \label{fig:qualitative}
\end{figure}

\begin{table}
  \caption{Translator ablation on FLUX.2-klein (Qwen3-0.6B + translator vs.\ the 4B encoder). Mem = encoder-path bf16 weights (shared 0.6B base ${\approx}1.4$\,GB + translator, params$\times$2\,B); the 4B row is the 4B encoder alone. Encode = text-encode latency (0.6B + translator), PyTorch/MPS, bf16. LPIPS/CLIPScore: end-to-end on 24 PartiPrompts vs.\ 4B at matched seed. $^\dagger$seed-variance floor (4B at two seeds). Attention rows timed on CPU (MPS SDPA fallback).}
  \label{tab:ablation}
  \small
  \setlength{\tabcolsep}{4.5pt}
  \begin{tabular}{lrrrrr}
    \toprule
    Translator & Params & Mem & Encode & LPIPS\,$\downarrow$ & CLIP\,$\uparrow$ \\
    \midrule
    4B (reference)   & 4000\,M & 8.0\,GB  & 435\,ms & 0.553$^\dagger$ & 0.328 \\
    \midrule
    d3072 (MLP)      & 187\,M & 1.77\,GB & 107\,ms & \textbf{0.514} & 0.306 \\
    d2048$\times$8   & 158\,M & 1.72\,GB & 105\,ms & 0.550 & 0.299 \\
    d2048$\times$7    & 98\,M  & 1.60\,GB & 99\,ms  & 0.564 & 0.295 \\
    d2048 wide       & 91\,M  & 1.58\,GB & 98\,ms  & 0.580 & 0.298 \\
    d2048            & 91\,M  & 1.58\,GB & 98\,ms  & 0.578 & 0.295 \\
    d2048 last       & 85\,M  & 1.57\,GB & 98\,ms  & 0.610 & 0.265 \\
    d1024            & 29\,M  & 1.46\,GB & 92\,ms  & 0.661 & 0.238 \\
    d1024$\times$1     & 16\,M  & 1.43\,GB & 90\,ms  & 0.660 & 0.238 \\
    d512             & 10\,M  & 1.42\,GB & 90\,ms  & 0.710 & 0.179 \\
    d512$\times$2      & 8\,M   & 1.42\,GB & 90\,ms  & 0.709 & 0.180 \\
    d512 2-lin.      & 6\,M   & 1.41\,GB & 89\,ms  & 0.712 & 0.185 \\
    d256             & 3.5\,M & 1.41\,GB & 89\,ms  & 0.717 & 0.166 \\
    \midrule
    attn d1536       & 112\,M & 1.63\,GB & 164\,ms & 0.707 & 0.184 \\
    attn d1024       & 75\,M  & 1.55\,GB & 149\,ms & 0.714 & 0.171 \\
    \bottomrule
  \end{tabular}
\end{table}

\section{Speed/Quality/Memory Recipe}
\label{sec:system}

We attack the speed/quality/memory triangle with two orthogonal tools applied to the denoising transformer, which dominates all three axes: it is the largest component by weight, the slowest by accumulative time, and the most sensitive to precision loss.

\paragraph{Quantization}
We use post-training quantization (PTQ) exclusively, applied directly to any pretrained checkpoint---in contrast to quantization-aware training (QAT), which can recover accuracy at very low bit-widths but requires training data and a fine-tuning loop. Quantization affects all three axes simultaneously: lowering weight precision reduces VRAM; activation quantization on FP8~\cite{micikevicius2022fp8formatsdeeplearning}/NVFP4~\cite{nvidia2025nvfp4} tensor cores cuts compute time; and quality is slightly affected relative to BF16, as measured in Table~\ref{tab:quant}. FP16 and FP8 remain close to BF16, while NVFP4 shows the largest visual distance (LPIPS 0.23, CLIP sim.\ 0.92) with the highest prompt-to-prompt variance, indicating that quality degradation at that precision is content-dependent; orthogonal 4-bit weight-quantization methods such as SVDQuant~\cite{li2024svdquant} could compose with our pipeline. Visual comparisons are provided in the supplemental material. On the RTX PRO 6000 Blackwell (Table~\ref{tab:quant}). Low-bit quantization cuts per-step latency from 105\,ms to 57\,ms ($1.84\times$), pushing TTFI below 150\,ms. Independent of the exact GPU SKU, the quantized transformer engine reduces its VRAM requirement from 8.4\,GiB (BF16) down to 3.3\,GiB for NVFP4 on any Blackwell GPU, with the Ada generation reaching 4.8\,GiB using FP8.

Beyond FP8 and NVFP4, FP16 is a worth\-while precision to evaluate alongside BF16 on GeForce hardware, since tensor core throughput for BF16 or FP16 with FP32 accumulation is just half of the peak FP16 throughput~\cite{blackwell2025arch}. For a compute-bound stage such as the denoising transformer, this can translate to significant speedups.

\begin{table}
  \caption{Transformer precision on RTX PRO 6000 Blackwell, $1024\times1024$, single step, 4B encoder. Weights/Act.\ = weight/peak-activation memory. Quality vs.\ BF16 on 24 PartiPrompts: CLIP image similarity and LPIPS~\cite{zhang2018lpips} (mean\,$\pm$\,std).}
  \label{tab:quant}
  \small\setlength{\tabcolsep}{4pt}
  \begin{tabular}{lrrrrcc}
    \toprule
    Prec. & Step  & Weights & Act.  & CLIP\,$\uparrow$ & LPIPS\,$\downarrow$ \\
    \midrule
    BF16  & 105\,ms & 7{,}9\,GiB & 510\,MiB & \multicolumn{2}{c}{ref.} \\
    FP16  & 108\,ms & 7{,}9\,GiB & 510\,MiB & $0.99\pm0.01$ & $0.02\pm0.02$ \\
    FP8   &  73\,ms & 4{,}3\,GiB & 462\,MiB & $0.96\pm0.05$ & $0.10\pm0.07$ \\
    NVFP4 &  57\,ms & 2{,}8\,GiB & 475\,MiB & $0.92\pm0.08$ & $0.23\pm0.09$ \\
    \bottomrule
  \end{tabular}
\end{table}

\paragraph{Weight streaming}
Weight streaming keeps only a resident subset of denoiser weights in VRAM and transfers the remainder from pinned RAM on demand, overlapping PCIe transfers with compute. Because only weight residency changes, outputs are bit-identical to a fully resident run and quality is untouched, unlike quantization. It also does not reduce overall memory pressure, since the host must still hold the full model in RAM; it is therefore a poor fit for unified-memory (UMA) systems where VRAM and RAM share one pool. On discrete GPUs it converts VRAM scarcity into a latency cost governed by the \emph{reservoir decay factor} $\lambda = L \cdot M / W$: $L$ the execution latency between successive matrix multiplies, $M$ the host--device PCIe rate, $W$ the layer weight size. When $\lambda \geq 1$ every transfer hides behind computation (zero added latency); when $\lambda < 1$ the reservoir drains faster than it refills and awaited weights add overhead $\Delta = (N{-}K)\cdot W/M$ ($N$ layers, $K \leq N$ fully overlapped). Crucially $M$ is fixed by the host platform---a high-end RTX~5090 and a mobile RTX~5070 share the same PCIe Gen\,5 slot---while compute throughput varies by an order of magnitude, so $\lambda$ is far closer to 1 on a slow, low-VRAM GPU than on a fast one: the device that needs streaming most is where it is cheapest.

Weight streaming reduces the fixed memory occupied by model weights from $N \cdot W$ down toward a smaller set of resident weights in GPU memory, but it introduces additional scratch memory that must remain on device. Now not only activation tensors are required as scratch space but weights for layer $i{+}1$ and $i$ to overlap computation. The minimum viable device memory target is therefore $T_{\min} = 2\cdot W_{\max} + A_{\max}$, where $W_{\max}$ is the largest single-layer weight tensor in the network and $A_{\max}$ is the peak activation tensor that must coexist with the weights during the forward pass. Streaming cannot reduce $T$ below $T_{\min}$ regardless of how aggressively blocks are evicted.

\begin{figure*}
  \centering
  \includegraphics[width=0.97\textwidth]{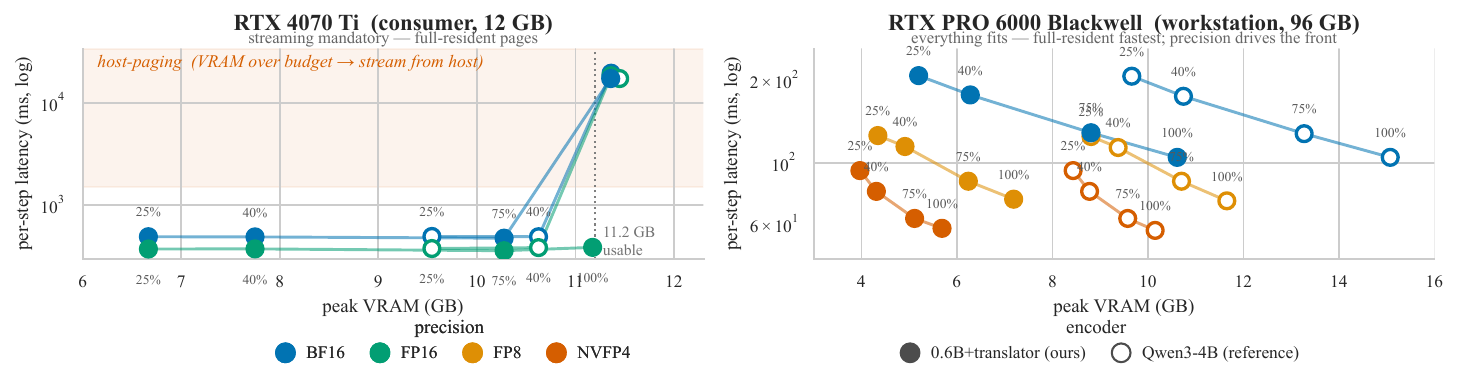}
  \caption{Cross-device sweep (4 steps, $1024^2$): peak VRAM vs.\ per-step latency. Label = resident-weight budget (100\% = full); line = one (encoder, precision) trajectory; filled = 0.6B+translator, open = 4B. \emph{RTX~4070~Ti (12\,GB):} above the ${\approx}11.2$\,GB budget the denoiser pages, latency jumps ${\sim}40\times$ (shaded). \emph{RTX~PRO~6000 (96\,GB):} all resident---full-resident is fastest and precision drives the front. BF16 exact, FP16 ${\geq}0.99$ vs BF16.}
  \label{fig:sweep}
\end{figure*}

Figure~\ref{fig:sweep} (left) sweeps all BF16 and FP16 configurations on the RTX~4070~Ti (12\,GB); the full numerical breakdown is in the supplemental material. With the 4B encoder, 75\,\% and full residency both page ($\lambda \ll 1$, ${\sim}40\times$ slowdown) because the encoder alone consumes ${\approx}8$\,GB, leaving too little headroom for system resources. Replacing it with the 0.6B+translator encoder opens up the budget: 25\,\% through 75\,\% residency all land in the $\lambda \approx 1$ regime with latency within 3\,\% of each other; only disabling streaming entirely crosses the paging cliff. At 25\,\% residency with the translator path, peak VRAM drops to 6.7\,GB---well under the 8\,GB tier---at only a 3\,\% step-time cost. FP16 is consistently 23--25\,\% faster than BF16 at every streaming level, confirming the GeForce throughput advantage discussed above. The right panel shows the RTX~PRO~6000~Blackwell, where the full model fits in VRAM: here streaming at low budgets (ws25, ws40) is \emph{slower} than full residency, because the fast Ada compute drains the reservoir faster than PCIe can refill it ($\lambda < 1$), illustrating that streaming is only beneficial when the model would otherwise not fit.

The two techniques compose freely: quantization compresses the weights that streaming moves, so applying both multiplies the VRAM reduction while accelerating step times significantly. Since PTQ engines ship as a single set of weights, different precision formats require no additional storage.

\section{Interactive On-Device Image Generation Editor}
\label{sec:editor}

We combine the translator, quantization, and offloading into an interactive image generation editor that runs entirely on-device, targeting sub-second TTFI on recent GPUs. The editor is a native C++ application on ONNX Runtime's TensorRT-RTX provider; each component's engine is compiled once and cached, so switching encoder, precision, or streaming budget at runtime needs no rebuild.

In the editor, encoder, precision, and weight-streaming residency are live controls with per-stage timings and GPU memory; even on a 12\,GB RTX~4070~Ti the 0.6B+translator path at half-resident weights renders $1024\times1024$ in about 3\,s with headroom to spare.

\paragraph{RTX PRO 6000 Blackwell measurements.} Table~\ref{tab:blackwell} sweeps the four configurations on an NVIDIA RTX PRO 6000 Blackwell \mbox{Workstation} Edition using TensorRT-RTX, generating a $1024\times1024$ image in four distilled steps. Replacing the 4B encoder with the 0.6B+translator (enc1) cuts VRAM from 15.1\,GB to 10.6\,GB with no change in per-step latency, since the encoder runs in 4\,ms versus 12\,ms and is not on the critical path. Quantization then drives both axes: FP8 halves the VRAM footprint to 7.2\,GB and reduces step time to 74\,ms; NVFP4 reaches 5.7\,GB and 57\,ms per step, bringing the full four-step end-to-end time to 0.27\,s and TTFI to ${\approx}100$\,ms.

\begin{table}
  \caption{On-device timings, NVIDIA RTX PRO 6000 Blackwell Workstation Edition, TensorRT-RTX, $1024\times1024$, 4 steps. TTFI = encode + one denoising step + VAE decode. VRAM = peak device memory across the full pipeline.}
  \label{tab:blackwell}
  \small
  \setlength{\tabcolsep}{4.5pt}
  \resizebox{\columnwidth}{!}{%
  \begin{tabular}{llrrrrrr}
    \toprule
    Encoder & Prec. & Encode & Step & Decode & TTFI & Total & VRAM \\
    \midrule
    4B      & BF16  & 12\,ms & 105\,ms & 41\,ms & 158\,ms & 0.47\,s & 15.1\,GB \\
    0.6B+tr & BF16  &  4\,ms & 105\,ms & 40\,ms & 149\,ms & 0.47\,s & 10.6\,GB \\
    0.6B+tr & FP8   &  4\,ms &  73\,ms & 41\,ms & 119\,ms & 0.34\,s &  7.2\,GB \\
    0.6B+tr & NVFP4 &  4\,ms &  57\,ms & 40\,ms & 102\,ms & 0.27\,s &  5.7\,GB \\
    \bottomrule
  \end{tabular}}
\end{table}

\section{Conclusion}
Text conditioning for diffusion does not require a large encoder as used in training and can rely on a small post-hoc translator recovering much of a 4B encoder's quality, using an embedder the device already hosts. Combined with quantization and offloading, this brings interactive diffusion to a much wider set of consumer GPUs.

\paragraph{Data and licensing.} FLUX.2-klein, both Qwen3 encoders, and PartiPrompts~\cite{yu2022parti} are Apache-2.0; all inference runs on-device. Code: \url{https://github.com/NVIDIA/din-deploy}.

\bibliographystyle{ACM-Reference-Format}
\bibliography{references}

@inproceedings{rombach2022ldm,
  title={High-Resolution Image Synthesis with Latent Diffusion Models},
  author={Rombach, Robin and Blattmann, Andreas and Lorenz, Dominik and Esser, Patrick and Ommer, Bj{\"o}rn},
  booktitle={IEEE/CVF Conference on Computer Vision and Pattern Recognition (CVPR)},
  year={2022}
}

@inproceedings{esser2024sd3,
  title={Scaling Rectified Flow Transformers for High-Resolution Image Synthesis},
  author={Esser, Patrick and Kulal, Sumith and Blattmann, Andreas and Entezari, Rahim and M{\"u}ller, Jonas and Saini, Harry and Levi, Yam and Lorenz, Dominik and Sauer, Axel and Boesel, Frederic and others},
  booktitle={International Conference on Machine Learning (ICML)},
  year={2024}
}

@misc{flux2024,
  title={{FLUX}},
  author={{Black Forest Labs}},
  year={2024},
  howpublished={\url{https://github.com/black-forest-labs/flux}}
}

@article{qwen3,
  title={Qwen3 Technical Report},
  author={{Qwen Team}},
  journal={arXiv preprint arXiv:2505.09388},
  year={2025}
}

@inproceedings{hessel2021clipscore,
  title={{CLIPScore}: A Reference-free Evaluation Metric for Image Captioning},
  author={Hessel, Jack and Holtzman, Ari and Forbes, Maxwell and Le Bras, Ronan and Choi, Yejin},
  booktitle={Conference on Empirical Methods in Natural Language Processing (EMNLP)},
  year={2021}
}

@inproceedings{zhang2018lpips,
  title={The Unreasonable Effectiveness of Deep Features as a Perceptual Metric},
  author={Zhang, Richard and Isola, Phillip and Efros, Alexei A. and Shechtman, Eli and Wang, Oliver},
  booktitle={IEEE/CVF Conference on Computer Vision and Pattern Recognition (CVPR)},
  year={2018}
}

@article{yu2022parti,
  title={Scaling Autoregressive Models for Content-Rich Text-to-Image Generation},
  author={Yu, Jiahui and Xu, Yuanzhong and Koh, Jing Yu and Luong, Thang and Baid, Gunjan and Wang, Zirui and Vasudevan, Vijay and Ku, Alexander and Yang, Yinfei and Ayan, Burcu Karagol and others},
  journal={Transactions on Machine Learning Research (TMLR)},
  year={2022}
}

@article{hinton2015distilling,
  title={Distilling the Knowledge in a Neural Network},
  author={Hinton, Geoffrey and Vinyals, Oriol and Dean, Jeff},
  journal={arXiv preprint arXiv:1503.02531},
  year={2015}
}

@article{hu2024ella,
  title={{ELLA}: Equip Diffusion Models with {LLM} for Enhanced Semantic Alignment},
  author={Hu, Xiwei and Wang, Rui and Fang, Yixiao and Fu, Bin and Cheng, Pei and Yu, Gang},
  journal={arXiv preprint arXiv:2403.05135},
  year={2024}
}

@inproceedings{zhao2024lavibridge,
  title={Bridging Different Language Models and Generative Vision Models for Text-to-Image Generation},
  author={Zhao, Shihao and Hao, Shaozhe and Zi, Bojia and Xu, Huaizhe and Wong, Kwan-Yee K.},
  booktitle={European Conference on Computer Vision (ECCV)},
  year={2024}
}

@misc{micikevicius2022fp8formatsdeeplearning,
  title={FP8 Formats for Deep Learning},
  author={Paulius Micikevicius and Dusan Stosic and Neil Burgess and Marius Cornea and Pradeep Dubey and Richard Grisenthwaite and Sangwon Ha and Alexander Heinecke and Patrick Judd and John Kamalu and Naveen Mellempudi and Stuart Oberman and Mohammad Shoeybi and Michael Siu and Hao Wu},
  year={2022},
  eprint={2209.05433},
  archivePrefix={arXiv},
  primaryClass={cs.LG},
  url={https://arxiv.org/abs/2209.05433}
}

@inproceedings{li2023snapfusion,
  title={{SnapFusion}: Text-to-Image Diffusion Model on Mobile Devices within Two Seconds},
  author={Li, Yanyu and Wang, Huan and Jin, Qing and Hu, Ju and Chemerys, Pavlo and Fu, Yun and Wang, Yanzhi and Tulyakov, Sergey and Ren, Jian},
  booktitle={Advances in Neural Information Processing Systems (NeurIPS)},
  year={2023}
}

@article{zhao2024mobilediffusion,
  title={{MobileDiffusion}: Instant Text-to-Image Generation on Mobile Devices},
  author={Zhao, Yang and Xu, Yanwu and Xiao, Zhisheng and Jia, Haolin and Hou, Tingbo},
  journal={arXiv preprint arXiv:2311.16567},
  year={2024}
}

@misc{nvidia2025nvfp4,
  title={{Introducing NVFP4 for Efficient and Accurate Low-Precision Inference}},
  author={{NVIDIA Corporation}},
  year={2025},
  howpublished={\url{https://developer.nvidia.com/blog/introducing-nvfp4-for-efficient-and-accurate-low-precision-inference/}}
}

@techreport{blackwell2025arch,
  title={{NVIDIA RTX Blackwell GPU Architecture}},
  author={{NVIDIA Corporation}},
  institution={NVIDIA Corporation},
  year={2025},
  url={https://images.nvidia.com/aem-dam/Solutions/geforce/blackwell/nvidia-rtx-blackwell-gpu-architecture.pdf}
}

@inproceedings{wang2025scalingdown,
  title={Scaling Down Text Encoders of Text-to-Image Diffusion Models},
  author={Wang, Lifu and Liu, Daqing and Liu, Xinchen and He, Xiaodong},
  booktitle={IEEE/CVF Conference on Computer Vision and Pattern Recognition (CVPR)},
  year={2025}
}

@inproceedings{li2024svdquant,
  title={{SVDQuant}: Absorbing Outliers by Low-Rank Components for 4-Bit Diffusion Models},
  author={Li, Muyang and Lin, Yujun and Zhang, Zhekai and Cai, Tianle and Li, Xiuyu and Guo, Junxian and Xie, Enze and Meng, Chenlin and Zhu, Jun-Yan and Han, Song},
  booktitle={International Conference on Learning Representations (ICLR)},
  year={2025}
}

\end{document}